%% file: main.tex
\documentclass{article}

\usepackage[final]{corl_2020}
\usepackage{latexsym}
\usepackage{booktabs}
\usepackage{amsmath}
\usepackage{xcolor,colortbl}
\usepackage{pifont}
\usepackage{multirow}
\usepackage{graphicx}
\usepackage{subcaption} 
\usepackage{url}

\newcommand{\datasetfull}{Robot Simultaneous Localization and Mapping with Natural Language}
\newcommand{\dataset}{RobotSlang}
\newcommand{\loctaskfull}{Localization from Dialog History}
\newcommand{\loctask}{LDH}
\newcommand{\navtaskfull}{Navigation from Dialog History}
\newcommand{\navtask}{NDH}

\newcommand{\dri}{\textsc{Driver}}
\newcommand{\com}{\textsc{Commander}}
\newcommand{\drishort}{\textsc{Dri}}
\newcommand{\comshort}{\textsc{Com}}

\newcommand{\topometricfull}{Topological Distance}
\newcommand{\topometric}{TD}

\newcommand{\numdialogs}{169}

\newcommand{\cblkmark}{\ding{51}}
\newcommand{\cmark}{\color{blue}{\ding{51}}}
\newcommand{\xmark}{\color{red}{\ding{55}}}
\newcommand{\good}[1]{\textcolor{blue}{\textbf{#1}}}
\newcommand{\bad}[1]{\textcolor{red}{\textbf{#1}}}
\definecolor{mypurple}{RGB}{150, 0, 255}
\newcommand{\neutral}[1]{\textcolor{mypurple}{\textbf{#1}}}
\definecolor{Gray}{gray}{0.95}
\newcolumntype{a}{>{\columncolor{Gray}}r}
\newcommand{\textquoteinline}[1]{\textit{#1}}

\title{The \dataset\ Benchmark: \\Dialog-guided Robot Localization and Navigation}

\author{
  Shurjo Banerjee$^1$ \qquad Jesse Thomason$^2$ \qquad Jason J. Corso$^1$ \\
  $^1$The University of Michigan \hspace{3em} $^2$University of Washington \\
  \texttt{\{shurjo, jjcorso\}@umich.edu} \hspace{1em} \texttt{jdtho@cs.washington.edu}
}

\begin{document}
\maketitle


\begin{abstract}
Autonomous robot systems for applications from search and rescue to assistive guidance should be able to engage in natural language dialog with people.
To study such cooperative communication, we introduce \datasetfull\ (\dataset), a benchmark of \numdialogs\ natural language dialogs between a human \dri\ controlling a robot and a human \com\ providing guidance towards navigation goals.
In each trial, the pair first cooperates to localize the robot on a global map visible to the \com, then the \dri\ follows \com\ instructions to move the robot to a sequence of target objects.
We introduce a \loctaskfull\ (\loctask) and a \navtaskfull\ (\navtask) task where a learned agent is given dialog and visual observations from the robot platform as input and must localize in the global map or navigate towards the next target object, respectively.
\dataset\ is comprised of nearly 5k utterances and over 1k minutes of robot camera and control streams.
We present an initial model for the \navtask\ task, and show that an agent trained in simulation can follow the \dataset\ dialog-based navigation instructions for controlling a physical robot platform.
Code and data are available at \url{https://umrobotslang.github.io/}.
\end{abstract}

\keywords{Benchmark, Robot, Visual Navigation, Natural Language, Dialog} 

\section{Introduction}
\label{sec:introduction}

\input{writing/01introduction}

\section{Related Work}
\label{sec:related_work}
\input{writing/02related_work}


\section{\datasetfull}
\label{sec:dataset}
\input{writing/03dataset}

\section{Tasks}
\label{sec:tasks}
\input{writing/04tasks}


\section{Experiments}
\label{sec:experiments}
\input{writing/05experiments}


\section{Conclusions and Future Work}
\label{sec:conclusion}
\input{writing/06conclusions}



\acknowledgments{The authors are supported in part by ARO grant (W911NF-16-1-0121) and by the US National Science Foundation National Robotics Initiative under Grants 1522904. Additionally, we would like to thank Dr. Jeffrey M. Siskind and Dr. Vikas Dhiman for their substantive discussions that were fundamental to RobotSlang's evolution. Finally, we would like to thank Matthew Dorrow for his help with administrating annotators and data collection.}


\bibliography{main}  


\clearpage
\section{Supplementary Material}
\input{writing/0Nsupp}

\end{document}

%% file: writing/01introduction.tex
Language is a natural medium for people to communicate with and direct robots.
Research on language-guided robots lets users specify high-level goals like \textquoteinline{Push the full barrel}~\cite{patki19a,arkin:ijrr20} and lower-level instructions like \textquoteinline{After the blue bale fly to the right towards the small white bush}~\cite{blukis:corl19,barrett2017driving}.
Meanwhile, commercial dialog-enabled smart assistants are imbued with expanded language capabilities, but have limited interaction with the real world.
Dialog-enabled robots combine these strengths, facilitating task completion~\cite{tellex:rss14}, task learning~\cite{chai:ijcai18,bullard:icra18}, and language learning~\cite{thomason:jair20}.

Two related skills are needed across potential robot applications, from search and rescue missions to assistive guidance in an office building: localization and navigation.
For a non-expert user, dialog facilitates both.
The user can ascertain \textit{where} the robot is by making requests like \textquoteinline{Describe your surroundings} and give instructions for where to go next, such as \textquoteinline{go toward the edge of the maze}.
There is an array of benchmarks for vision-and-language navigation (VLN), where an agent learns to follow such language directions in simulated environments~\cite{chen:aaai11,anderson:cvpr18,marge:naacl19,chen:cvpr19}.

Models optimizing performance on simulation-only benchmarks do not consider physical robot navigation limitations, employing unrealistic strategies like beam search~\cite{ke:cvpr19}, repeatedly creating panoramas from egocentric cameras~\cite{fried:nips18}, or assuming pre-exploration of the environment~\cite{tan:naacl19}.
Where simulation allows the collection of such large-scale benchmarks, efforts in learning-based robotics are limited by expensive data collection and require more robust, sample-efficient models.

\begin{figure}[t]
    \centering
    \includegraphics[width=1.\linewidth]{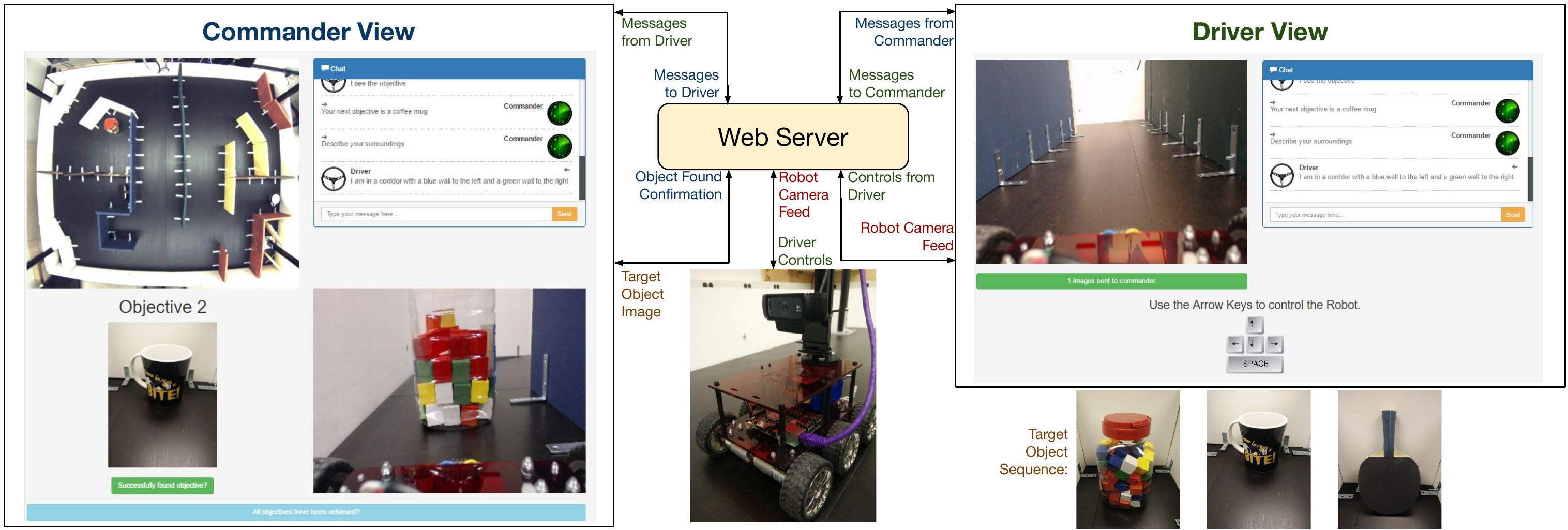}
    \caption{The \com\ and \dri\ communicate through a web application.
        The \com\ sees a static map and an image of the next target object (left).
        The \dri\ sees the first-person camera feed from the robot (a small rover controlled by a raspberry pi with an attached HD webcam) and can send the robot movement commands (right).
        Participants cooperate through a chat interface to guide the robot to a sequence of three target objects (bottom right).
        We synchronize, capture, and annotate the raw dialog text, sensor observations, and map data to create the \dataset\ benchmark.
    }
    \label{fig:schema}
\end{figure}

To address this gap, we introduce \datasetfull\ (\dataset), a benchmark of human-human cooperative trials for controlling a physical robot to visit object goals while communicating in natural language (Figure~\ref{fig:schema}).
\dataset\ data is collected on a physical robot platform and is directly applicable for training a language-guided robot to follow human language instructions, which can otherwise require careful transfer techniques~\cite{anderson_vlnpano2real_2020}.

%% file: writing/02related_work.tex
A learning agent reacting to human language instructions should be \textit{embodied} in the real world and be able to carry on \textit{dialog} with human users~\cite{bisk:arxiv20}.
To this end, \dataset{} combines aspects of two research efforts: language-guided virtual agents in simulation and situated, human-robot dialog for guiding robot behavior.
Table~\ref{tab:rw_comparison} summarizes the key differences between \dataset\ and comparable benchmarks across the language, vision, and robotics communities.
\dataset\ provides a resource for studying how humans use language to cooperatively control a physical robot.
As with initiatives like the Duckietown~\cite{paull2017duckietown} and MuSHR~\cite{srinivasa2019mushr}, our physical robot setup can be recreated in other labs to make use of \dataset\ data for training and evaluation.

\paragraph{Language-Guided Virtual Agents.}
Given a natural language instruction, an agent attempts to \textit{ground}~\cite{harnad:phys90} this language to its visual surroundings.
Early environments used simple texture renderings~\cite{macmahon:aaai06,chen:aaai11}, but advances in simulation and scene capture have lead to photorealistic indoor~\cite{anderson:cvpr18} and outdoor~\cite{chen:cvpr19} spaces for VLN and general video understanding~\cite{zhou:cvpr19}.
Some benchmarks include object manipulation and state changes, creating a more task-oriented setting~\cite{shridhar:cvpr20}.

Moving beyond static instructions, some benchmarks are created via human-human dialog, where a \dri\ moves the agent and asks questions that a \com\ answers.
By enabling an agent to ask questions and gather more information during navigation~\cite{nguyen:cvpr19,nguyen:emnlp19}, ambiguous and underspecified commands can be clarified by a human interlocutor~\cite{devries:arxiv18,marge:naacl19,thomason:corl19}.
Some of these benchmarks require global localization~\cite{devries:arxiv18}, where the \com\ does not initially know enough about the location of the \dri\ to provide instructions.
These benchmarks are all limited to simulations, while deployed robots operate in noisier, real-world environments.
Our \dataset{} benchmark, by contrast, was gathered by pairing a human \dri\ controlling a physical robot and asking questions of a human \com, where the pair need to perform cooperative global localization while carrying out a navigation task to multiple object targets.

\begin{table}[t]
\centering
\begin{small}
\begin{tabular}{lcccccc}
    \textbf{Comparable} & \textbf{Agent} & \textbf{Localization} & \textbf{Navigation} & \textbf{Questions} & \textbf{Human} & \textbf{Sensor} \\
    \textbf{Efforts} & \textbf{Type} & \textbf{Task} & \textbf{Task} & \textbf{Allowed} & \textbf{Answers} & \textbf{Observations} \\
    \toprule
    MARCO \cite{macmahon:aaai06,chen:aaai11} & \bad{Virtual} & \xmark & \cmark & \xmark & - & \bad{Render} \\
    VLN \cite{anderson:cvpr18,chen:cvpr19} & \bad{Virtual} & \xmark & \cmark & \xmark & - & \neutral{Photoreal} \\
    DRIF \cite{misra:emnlp18,blukis:corl19} & \good{Physical} & \xmark & \cmark & \xmark & - & \good{Camera RGB} \\
    DUIL \cite{barrett2017driving} & \good{Physical} & \xmark & \cmark & \xmark & - & \bad{Full Map} \\
    VLNA \cite{nguyen:cvpr19,nguyen:emnlp19} & \bad{Virtual} & \xmark & \cmark & \cmark & \xmark & \neutral{Photoreal} \\
    MRDwH \cite{marge:naacl19} & \bad{Virtual} & \xmark & \cmark & \cmark & \cmark & \bad{Render} \\
    CVDN\cite{thomason:corl19} & \bad{Virtual} & \xmark & \cmark & \cmark & \cmark & \neutral{Photoreal} \\
    Talk the Walk \cite{devries:arxiv18} & \bad{Virtual} & \cmark & \cmark & \cmark & \cmark & \neutral{Photoreal} \\
    \midrule
    \dataset{} & \good{Physical} & \cmark & \cmark & \cmark & \cmark & \good{Camera RGB} \\
    \bottomrule \\
\end{tabular}
\end{small}
\caption{
Compared to existing efforts involving vision and language input for controlling navigation agents, \dataset{} is the first to be built from human-human dialogs between a \com\ and \dri\ piloting a physical robot.
\dataset{} also involves both a global localization and navigation task, where many previous efforts involve only navigation.
}
\vspace{-6mm}
\label{tab:rw_comparison}
\end{table}

\paragraph{Language-Guided Physical Agents.}
Physical robots engaged in task-oriented behaviors with human collaborators benefit from a natural language interface~\cite{tellex:arcras:20}.
For example, human language commands can be combined with visual sensory input~\cite{misra:emnlp18} to learn a controller for a quadcopter platform in an end-to-end fashion~\cite{blukis:corl18, blukis:corl19}.
Using natural language and mixed reality can allow users to control quadcopters with little training data~\cite{huang19}.

Collaborative dialog can enable a robot to acquire new skills~\cite{chai:ijcai18} and refine perceptual understanding~\cite{thomason:jair20} through language interaction.
However, generating grounded language requests is often achieved through carefully controlled dialog managers~\cite{danas19errorrec} and generation semantics~\cite{tellex:rss14}.
Learning to generate language from grounded, human-human dialogs~\cite{devries:arxiv18,roman:arxiv20} such as those in the \dataset\ benchmark is an ambitious alternative for future work.

%% file: writing/03dataset.tex
In \datasetfull\ (\dataset), humans take on the roles of \com\ and \dri\ to guide a robot through a 2.5 by 2 meter tabletop maze to visit a series of three target landmark objects (Figure~\ref{fig:schema}).
In total, \dataset\ is comprised of 169 trials with associated dialogs and robot sensor streams carried out in 116 unique mazes.

Information and control asymmetry drives natural language communication to finish tasks, and a leaderboard was used to rank human-human teams and encourage efficient task completion (Section~\ref{ssec:collection}).
The \com\ viewed a top-down, static map of the environment that included the landmarks to be visited, but could not see the robot.
The \dri\ had access to the robot's sensor streams, including its front-facing camera, but did not have access to a global map.
The participants communicated only over a text chat interface.
First, the \dri\ needed to provide information about their surroundings to the \com\ to perform \loctaskfull\ (\loctask).
Then, the \com\ provided directions to the \dri.
The \dri\ could ask questions throughout.
Previous dialog-based navigation work forced strict turn taking and involved only one target per dialog~\cite{devries:arxiv18,thomason:corl19}.
In \dataset, communication is at-will, and each trial involves visiting a sequence of three target objects.
These differences lead to complex dialog phenomena (Section~\ref{ssec:analysis}).

\subsection{Data Collection}
\label{ssec:collection}

\begin{table}[t]
\centering
\begin{small}
\begin{tabular}{p{1.8cm}rp{9.2cm}}
    & \textbf{Freq (\%)} & \textbf{Examples} \\
    \toprule
    Needs History & $53.0$ & \comshort: Go back to {\color{blue}your initial position} \\
    \cmidrule{3-3}
    & & \comshort: now, go back to the location {\color{blue}where you found the jar at first} \\
    \cmidrule{3-3}
    & & \comshort: go back to the location {\color{blue}where you turn right first} \\
    \midrule
    Communication & $26.5$ & \comshort: Turn right before you go straight \\
    Repair & & \drishort: {\color{blue}what?} $|$ {\color{blue}Turn right where?} \\
    \cmidrule{3-3}
    & & \comshort: {\color{blue}are you sure} the blue walls are on your right? \\
    & & \drishort: left $|$ sorry \\
    \midrule
    Re-localization & $14.7$ & \comshort: i asked you to turn right at the brown wall, and take the right at the \phantom{\comshort: }white edge of the maxe \\
    Repair & & \comshort: {\color{blue}where are you right now?} \\
    \cmidrule{3-3}
    & & \drishort: after the red wall or before the red wall? \\
    & & \comshort: before $|$ {\color{blue}Give me your position} \\
    \midrule
    Surrender & $11.8$ & \comshort: your third objective is the ping-pong-paddle \\
    Control & &  \comshort: {\color{blue}can you go there on your own?} \\
    & & \drishort: yes \\
    \cmidrule{3-3}
    & & \comshort: ok {\color{blue}let me know} if you get lost [end of dialog] \\
    \midrule
    Perceptual & $\phantom{0}2.9$ & \comshort: then move forward and make a right uturn around the orange walls \\
    Differences & & \drishort: {\color{blue}it is red} \\
    & & \comshort: correct, sorry I keep forgetting \\
    \bottomrule \\
\end{tabular}
\end{small}
\caption{
Dialog phenomena we annotated, along with the estimated frequency and examples from a 34 dialogs randomly sampled for annotation (20\% of the total trials).
We use the symbol $|$ to separate short, sequential messages sent from the same speaker.
}
\vspace{-8mm}
\label{tab:analysis}
\end{table}

Each dialog in \dataset\ facilitated control of a small robotic car through a tabletop maze to visit a sequence of landmarks.\footnote{The infrastructure to collect this data, including software and details of robot hardware and materials to construct physical mazes, is available at \url{https://umrobotslang.github.io/}.}
Throughout each trial, the robot collected front-facing camera stream data.
The \com\ and \dri, although physically separated in different rooms, were connected via a web interface, shown in Figure~\ref{fig:schema}.
The \dri\ viewed a real-time stream from the robot camera, while the \com\ viewed a static, top-down RGB image of the maze and a photo of the next landmark object to visit.
Team members communicated via a text chat interface.
A central server hosted the web application and saved synchronized sensor and text dialog data streams.\footnote{See Supplementary Material for trial videos and details about the web application and tabletop maze setup.}

Sixteen participants were recruited to create \dataset.
Participants were not familiar with the robot platform and were students from the university at which this research was conducted.
The participants formed 41 teams, with each participant involved in 5 teams and 21 trials on average.
To motivate participant teams to do well on the task, each trial was scored, and teams competed on a leaderboard.
Teams were scored by overall time taken to visit all landmarks, with a small penalty for each motion command the \dri\ sent to the robot.
The same maze was never presented to a team twice, and each team performed an average of 4 trials, resulting in 169 total trials.
Dialogs lasted an average of six and a half minutes.

\subsection{Dialog Data Analysis}
\label{ssec:analysis}

The dialogs in \dataset\ are long.
An average of $28$ messages was sent per dialog, and dialogs averaged $200$ words.
By contrast, dialogs in the closely related, simulation-only CVDN task~\cite{thomason:corl19} contain an average of only $81.6$ words per dialog, less than half the verbosity in our trial data.

Table~\ref{tab:analysis} gives examples of dialog snippets from \dataset\ that exhibit complex phenomena.
Because each trial involves visiting a sequence of objects, over 50\% of dialogs involve \com{}s referring to past locations in their instructions.
Successful modeling may require explicit memory structures.
Humans make mistakes, both when giving instructions and when following them, but communication also facilitates repairing those mistakes.
Over 25\% of the dialogs in \dataset\ exhibit mistakes and repair.
Some of these mistakes require a new localization step, with the \com\ requesting location information again during navigation in 15\% of dialogs.
However, Humans also trust one another to follow ``commonsense'' routes and directions, with \com{}s surrendering control to \dri{}s in around 10\% of dialogs.
Finally, because the \com\ and \dri\ see different views of the environment, participants also need to mediate their perceptual differences explicitly in some dialogs, as studied in human-robot communication~\cite{liu:aaai15}.

\subsection{Replay-Simulation Environment}
\label{ssec:simulator}

\begin{figure}[t]
    \centering
    \includegraphics[width=\linewidth]{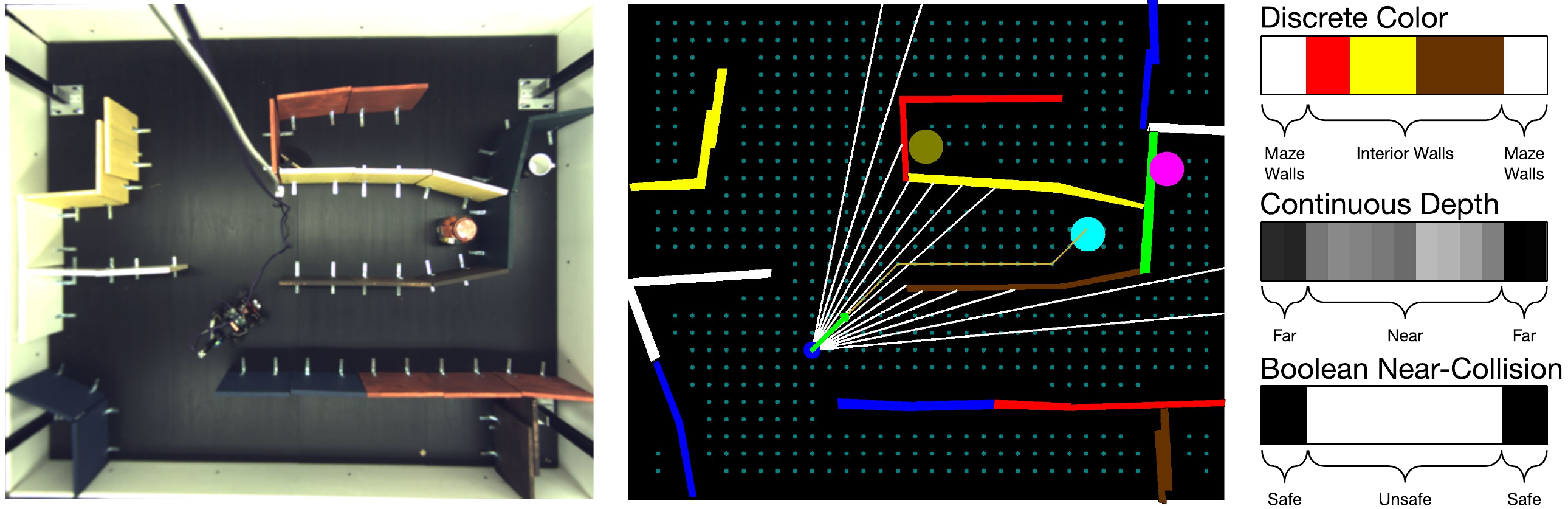} 
    \caption{\textbf{Left:} The real-world, top down view of a \dataset\ maze.
    \textbf{Center:} The replay-simulation view of the same maze, with discretized navigable points and vision based on ray tracing.
    The three objects are are given unique colors to differentiate them from the rest of the maze, and rays are shown here in white.
    \textbf{Right:} Sample visual input in the replay simulator, showing the colors, depths, and collision possibilities for 13 rays out of the front-facing camera.
    }
    \label{fig:replay_sim}
\end{figure}

In lieu of training on the physical tabletop environment, we design a custom simulator (Figure~\ref{fig:replay_sim}).
There are two aspects of the real world we simplify when creating the replay simulation environment: image observations and robot position.
We annotate trial maps as occupancy grids where grid values represent corresponding, discretized pixel colors from maze walls. 
Grid points are 0.07 meters apart.
The simulated agent travels between adjacent nodes using \texttt{forward} actions.
The agent can change its heading with \texttt{left} and \texttt{right} actions of 45 degrees each. 
We precompute all shortest paths between navigable points for use during model training (Section~\ref{ssec:ndh}).
For training, we treat the shortest path between the robot's starting position and each target visitation object in turn as the human navigation path.
We use the Floyd-Warshall algorithm~\cite{floyd1962algorithm} for shortest path planning.

We represent front-facing camera observations as a set of rays returning color and depth information.
In particular, the agent's observation at every timestep consists of, for each ray, a discrete wall color, a distance to the wall, and whether the wall is too close to move towards (Figure~\ref{fig:replay_sim}).
Thirteen equally spaced rays are cast from the front of the agent, encompassing a field-of-view of 78 degrees---the same as the physical robot.
We use Brensenham's line algorithm~\cite{bresenham1965algorithm} to perform ray tracing. 

The replay simulation allows training and testing models for tasks derived from \dataset.
These models may be transferable to the physical platform for either fine-tuning with real-world camera input or preprocessed image observations using the simulator-familiar ray tracing method.\footnote{Due to COVID-19, we were not able to validate this transfer on the physical platform. As noted in the Supplementary Material, we validate the effectiveness of our ray-tracing measurements by utilizing them to successfully localize the robot in the \dataset trials from the \com \ perspective.}

%% file: writing/04tasks.tex
\begin{table}[t]
\centering
\begin{tabular}{lrrrrr}
    \textbf{Fold} & \textbf{\# Mazes} & \textbf{\# Trials} & \textbf{\# \loctask} & \textbf{\# \navtask} & \textbf{\% Dataset by Trial} \\
    \toprule
    Train & 79 & 120 & 120 & 360 & 69 \\
    Val & 19 & 26 & 26 & 78 & 15 \\
    Test & 20 & 28 & 28 & 84 & 16 \\
    \bottomrule \\
\end{tabular}
\caption{
    Fold summaries in the \dataset\ benchmark, including the number of mazes, trials, and task instances.
    We split by mazes, such that no two folds contain trials with the same maze.
}
\label{tab:fold_stats}
\end{table}

\begin{figure*}[t]
    \centering
    \includegraphics[width=1\textwidth]{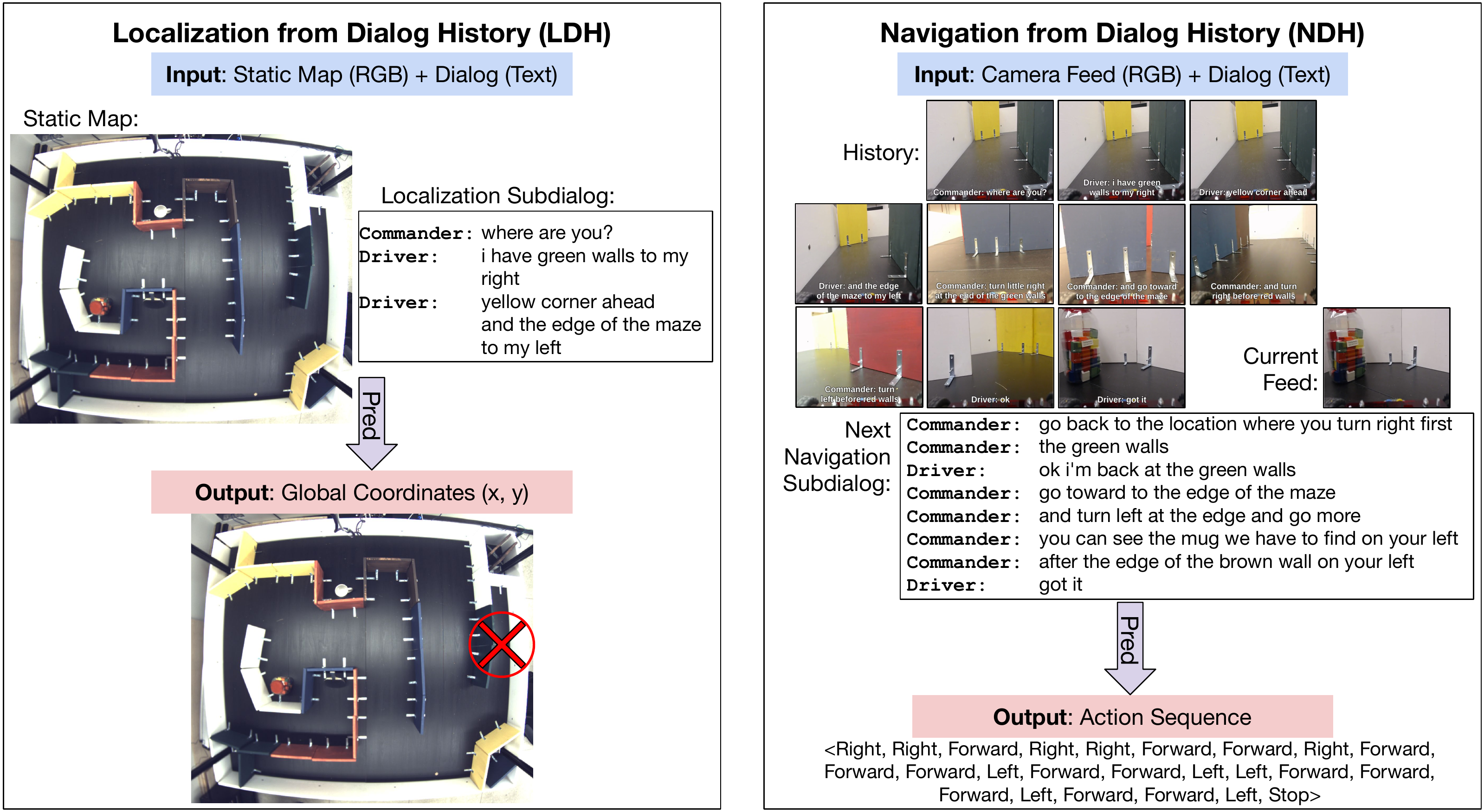}
    \caption{Instances of the \loctask\ task (left) and \navtask\ task (right).
        These tasks represent key skills of the \com\ and \dri: localization and navigation given natural language context.
        In both tasks, models have access to the dialog history so far as input.
        In \loctask, models also see the global static map, and must predict where the robot is based on the dialog.
        In \navtask, models also see the navigation history so far and the dialog snippet that guided the human-human pair to the next object, and must predict the navigation actions to reach that object.}
    \label{fig:task_examples}
\end{figure*}

\dataset\ provides a resource for approaching numerous problems in human-robot communication, from query generation---learning how to ask good questions---to language-driven dynamics prediction---learning to predict the \dri's actions in response to language.
While some work attempts to create agents that mimic both the \com\ and \dri, these methods are applied exclusively on data gathered in simulation and without regard to the need for sample efficiency with physical limitations on deployment~\cite{devries:arxiv18,roman:arxiv20}.
We instead focus on two core problems for any physical robot collaborating with a human giving language directions: \loctaskfull\ (\loctask) and \navtaskfull (\navtask).
We split the data into training, validation, and test folds (Table~\ref{tab:fold_stats}).
Figure~\ref{fig:task_examples} give example \loctask\ and \navtask\ instances.

\paragraph{\loctaskfull\ (\loctask)}

We create a benchmark for learning models that perform \loctaskfull, emulating the human \com.
We create one \loctask\ instance per trial.
For each trial, we annotated when initial localization was complete, before navigation began.
A model receives as input the dialog history between the \com\ and \dri\ from the trial start to that hand-annotated end of localization.
Given this information, the model must predict the location of the robot on the global map, just as the human \com\ does before giving navigation instructions.

As exhibited in Table~\ref{tab:analysis}, over 10\% of trials require explicitly re-localizing when participants realize they have mismatched assumptions.
In general, localization continues to happen in ``soft'' ways through the dialog, with the \com\ and \dri\ offering frequent sanity checks like \textquoteinline{You should then have blue on your left and white in front of you} and \textquoteinline{i have red walls ahead}.
Thus, models may need to explicitly continue performing localization after this initial \loctask\ step, which we leave for future work.
After initial localization, navigation can begin.

\paragraph{\navtaskfull\ (\navtask)}

We create a benchmark for learning models that perform \navtaskfull, emulating the human \dri.
For each trial, we create an \navtask\ instance for each of the three objects to be visited.
For each \navtask\ instance, the robot begins at the trial initial position or in front of the last object visited, and must navigate to the next object in the visitation sequence.
A model receives as input the robot navigation history, the dialog history between the \com\ and the \dri\ so far, and the dialog that guided the \dri\ from the starting location to the goal location.
Given this information, the model must predict the controller actions taken by the \dri\ to move the robot according to the given language instructions.
Actions are discretized to: \texttt{forward} (0.07 meters), \texttt{left} (45 degrees), \texttt{right} (45 degrees), and \texttt{stop}.
We evaluate \navtask\ performance based on the final position $(\hat{x}, \hat{y})$ of the robot.

\paragraph{Metrics.}

Both the \loctask\ and \navtask\ tasks involve predicting a location $(\hat{x}, \hat{y})$ on the map, either directly for localizing the robot or indirectly as the result of a sequence of navigation actions.
We evaluate performance with a \topometricfull\ (\topometric) metric that measures the shortest navigable path between the true position $(x^*, y^*)$ and the predicted position $(\hat{x}, \hat{y})$.
Concretely, $\topometric=|\text{shortestpath}((\hat{x}, \hat{y}), (x^*, y^*))|$.
To analyze how much error comes form predicting when to \texttt{stop}, we also measure agent performance under \textit{Oracle Stopping}~\cite{misra2017mapping}, how well the agent would have done if it had stopped at the optimal time along its predicted path $P$.
Oracle stopping distance O\topometric\ is calculated as $O\topometric=\min_{(x, y)\in P}|\text{shortestpath}((x, y), (x^*, y^*))|$ for $P$ the predicted path.

%% file: writing/05experiments.tex
\begin{figure*}[t]
    \centering
    \includegraphics[width=1\textwidth]{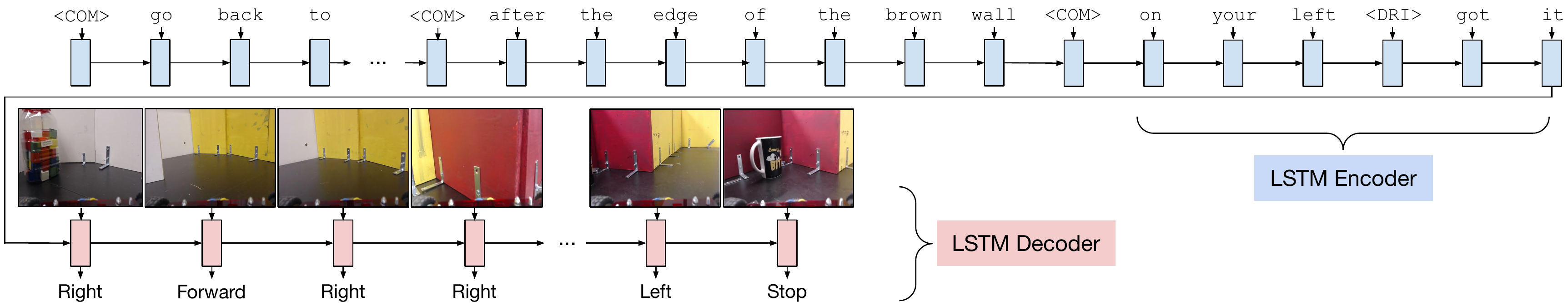}
    \caption{Our initial Sequence-to-Sequence model for the \navtask\ task.
    An LSTM encoder takes as input sequence of GloVe-embedded language tokens representing the dialog history.
    The encoder LSTM initializes the hidden state of an LSTM decoder, which takes in an encoded visual observation from the robot's  (or replay simulation agent's) front-facing camera and predicts an action.}
    \label{fig:model}
\end{figure*}

We evaluate human performance on the \loctask\ task and create an initial, learned model for the \navtask\ task which we evaluate in the replay simulation environment.
We find that humans perform well on the \loctask\ from reading dialog histories.
Our initial model for the \navtask\ outperforms simple baselines and ablations, indicating that the \dataset\ can be used to learn language navigation policies.

\subsection{\loctaskfull.}

The \loctask\ task involves making a discrete prediction for the location of the robot given the \com's static map view and the dialog history.
However, human \com{}s do not necessarily need the precise location of the robot, and may give instructions that are valid for a distribution around their belief of where the robot \textit{probably} is.

To assess the difference between humans' ability to make these discrete predictions and the physical locations of the robots, we conducted a human study with 8 participants who were not involved in the collection of \dataset.
Human predictions for the robot location were, on average over all \loctask\ instances, only $0.446$ meters away from the true robot center by \topometric.
This performance establishes an upper bound for future modeling attempts.
We note that while humans are not perfect at pinpointing the robot's location, their guesses are often within two robot body lengths of the true center, since the robot is about $0.254$ meters long.

\subsection{\navtaskfull.}
\label{ssec:ndh}

Following closely related work in simulation-only, human-human dialogs for navigation~\cite{thomason:corl19}, we focus our initial modeling efforts primarily on the \navtask\ task.
Given a dialog between a human \com\ and \dri, the robot agent must infer the next navigation actions to take, using the \dri's next actions as supervision during training.
Because the \dataset{} data was gathered on a physical robot platform, the \dataset{} benchmark can be trained and evaluated in both our replay simulation (Section~\ref{ssec:simulator}) and in the real world.

\paragraph{Initial Sequence-to-Sequence Model.}

We develop an initial, Sequence-to-Sequence (S2S) model, summarized in Figure~\ref{fig:model}.
This model mirrors initial S2S models used to evaluate prior work in vision-and-language navigation~\cite{anderson:cvpr18,thomason:corl19}.
The dialog history is tokenized and words are embedded using GloVe-300~\cite{pennington2014glove}.
Special tokens, \texttt{<COM>} and \texttt{<DRI>}, with random initial embeddings, are added to the sequence to indicate the speaker.
An LSTM~\cite{hochreiter1997long} model is used to encode this token sequence, and its final hidden state initializes an LSTM decoder model.
The LSTM decoder observes an image $I$ from the robot's front-facing camera, and predicts an action $\hat{a}$.
Under a \textit{teacher forcing} curriculum, at training time the ground truth action $a*$ from the human trial is taken.
Under a \textit{student forcing} curriculum, at training time an action $\hat{a}$ is taken after sampling from the predicted logits~\cite{anderson:cvpr18}.
All parameters are learned end-to-end by minimizing a cross entropy loss between predicted actions and the true action.
At inference time, predicted action $\hat{a}$ is taken in the simulation environment.\footnote{We build on the codebase infrastructure of prior work~\cite{thomason:corl19} for training and evaluation in simulation.}
We train and evaluate \navtask\ models using the replay simulation environment, approximating camera image observations with simulated ray tracing (Section~\ref{ssec:simulator}).

\paragraph{Results.}

\begin{table}[t]
    \centering
    \begin{tabular}{lccarar}
    & \multicolumn{2}{c}{\textbf{Inputs}} & \multicolumn{2}{c}{\textbf{Validation \topometric\ ($\pmb{\downarrow}$)}} & \multicolumn{2}{c}{\textbf{Test \topometric\ ($\pmb{\downarrow}$)}} \\ 
    \textbf{Training} & Vis & Lang & True & Oracle & True & Oracle \\ 
    \toprule
    Teacher & \cblkmark & & 3.77 $\pm$ 1.28 & 3.27 $\pm$ 1.21 & 3.67 $\pm$ 1.56 & 3.16 $\pm$ 1.50 \\ 
    Teacher & & \cblkmark & 4.01 $\pm$ 1.09 & 3.86 $\pm$ 1.06 & 4.03 $\pm$ 1.35 & 3.91 $\pm$ 1.31 \\ 
    Teacher & \cblkmark & \cblkmark & $^{*L}$3.68 $\pm$ 1.28 & 3.20 $\pm$ 1.27 & 3.76 $\pm$ 1.61 & 3.24 $\pm$ 1.51 \\
    \midrule
    Student & \cblkmark & & 3.42 $\pm$ 1.06 & 3.26 $\pm$ 1.11 & 3.32 $\pm$ 1.12 & 3.17 $\pm$ 1.10 \\ 
    Student & & \cblkmark & 4.04 $\pm$ 1.04 & 4.00 $\pm$ 1.03 & 4.10 $\pm$ 1.27 & 4.04 $\pm$ 1.26 \\
    Student & \cblkmark & \cblkmark & $^{*V,L}$3.27 $\pm$ 1.12 & 3.12 $\pm$ 1.13 & $^{*L}$3.38 $\pm$ 1.15 & 3.20 $\pm$ 1.14 \\
    \midrule
    \multicolumn{3}{l}{\texttt{Random Action}} & 4.32 $\pm$ 1.02 & 4.31 $\pm$ 1.02 & 4.31 $\pm$ 1.28 & 4.30 $\pm$ 1.28 \\ 
    \multicolumn{3}{l}{\texttt{Immediate Stop}} & 4.31 $\pm$ 1.02 & 4.31 $\pm$ 1.02 & 4.33 $\pm$ 1.28 & 4.33 $\pm$ 1.28 \\
    \bottomrule \\
    \end{tabular}
    \caption{Sequence-to-Sequence model results on the \navtask\ task under \textit{student-} versus \textit{teacher-} forcing, as well as under unimodal ablations~\cite{thomason:naacl19}.
    The \topometricfull\ (\topometric) is the primary measure of performance---the length of the shortest path between the predicted final location for the robot and the true location.
    We also show performance under \textit{Oracle Stopping}---the shortest distance achieved along the inferred navigation path.
    All metrics are in meters, and for all metrics lower is better.
    $^{*V,L}$ indicate when a model with access to both vision and language input statistically significantly outperforms the corresponding \textit{V}ision-only or \textit{L}anguage-only ablation.
    }
    \label{tab:results}
\end{table}
Table~\ref{tab:results} summarizes our results on the \navtask\ task.
Our initial, sequence to sequence model outperforms baselines as well as unimodal ablations.
Consistent with prior work~\cite{anderson:cvpr18}, training with a student forcing regime makes the full model significantly\footnote{See the Supplementary Material for details on the statistical comparison between models.} more robust to errors at inference time.
The random baseline selects an action (except \texttt{stop}) at random up to a maximum number of steps, while the immediate stop baseline selects \texttt{stop} as the first action.
These baselines provide context for the average distance agents need to cover to reach target objects in each \navtask\ instance.
In a unimodal ablation, the model is trained and tested with one modality (language or vision) empty.
Our initial model does not exhibit substantial unimodal bias pathology~\cite{thomason:naacl19}.
The test fold performance of the vision-only model is not statistically significantly different from the full model.

%% file: writing/06conclusions.tex
We introduce \datasetfull\ (\dataset), a dataset of human-human dialogs for controlling a physical robot, and create associated localization and navigation task benchmarks.
We present an initial model for \navtaskfull\ that improves over simple baselines, demonstrating that language-driven instruction following can be achieved from real-robot, human-human dialog data.

\paragraph{Limitations and Future Work}
While \dataset\ data was collected on a physical robot platform, our \navtaskfull\ evaluations were limited to our replay simulation environment.
In the future, we will evaluate agents trained in the replay simulation on the physical robot platform.
We are also interested in the possibilities of training on large scale vision-and-language navigation benchmarks and fine-tuning on \dataset\ as in-domain data for physical robot control.
While the underlying visual and action data distributions differ, similarities in natural language can be leveraged to create more robust models.

Currently, we focus only on localization and navigation tasks, but dialog can involve training two agents to entirely fill the roles of \com\ and \dri\, as tried in simulation-only VLN datasets~\cite{devries:arxiv18,roman:arxiv20}.
Our initial model for \navtask\ does not utilize explicit memory structures or historical visual features, but our analysis of dialog phenomena revealed that both are needed in over 50\% of dialogs to resolve commands like \textquoteinline{Go back to your initial position}.
Further modeling attempts on the \dataset\ navigation task can make use of historical sensory context beyond language.
We are also interested in exploring semi-supervised reinforcement learning for model training, which has been been successfully applied to multi-modal navigation tasks \cite{goyal2019using,dhiman2018critical}.

%% file: writing/0Nsupp.tex
\subsection{Full Example Dialogs from \dataset\ Trials}

\begin{figure*}[t]
    \centering
    \includegraphics[width=0.5\textwidth]{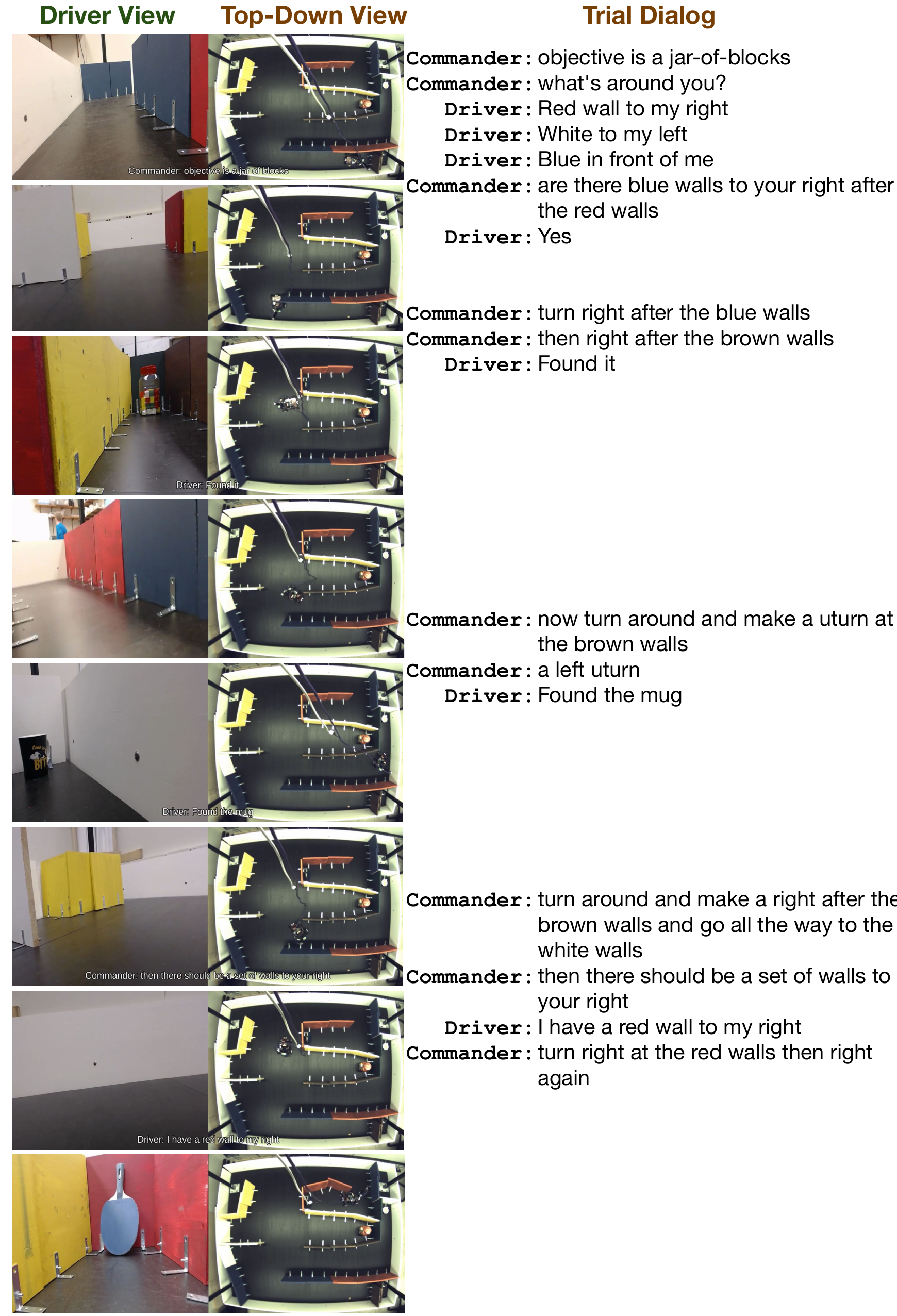}
    \caption{An example \dataset\ trial.
    The \com\ and \dri\ first work together to localize the robot, then the \com\ guides the \dri\ towards the target objects in sequence: a jar of blocks, a mug, and a paddle.
    The \dri\ sees the front-facing camera feed from the car and can control it, while the \com\ sees a static, top-down map.
    For visualization purposes, we show the live top-down recording view with the car, but the \com\ view during data collection was static.
    The dialog exchange between the \com\ and \dri\ was conducted over a text chat interface, and is shown here roughly aligned with frames from the recorded videos.
    }
    \label{fig:trial_178}
\end{figure*}

\begin{figure*}[t]
    \centering
    \includegraphics[width=1\textwidth]{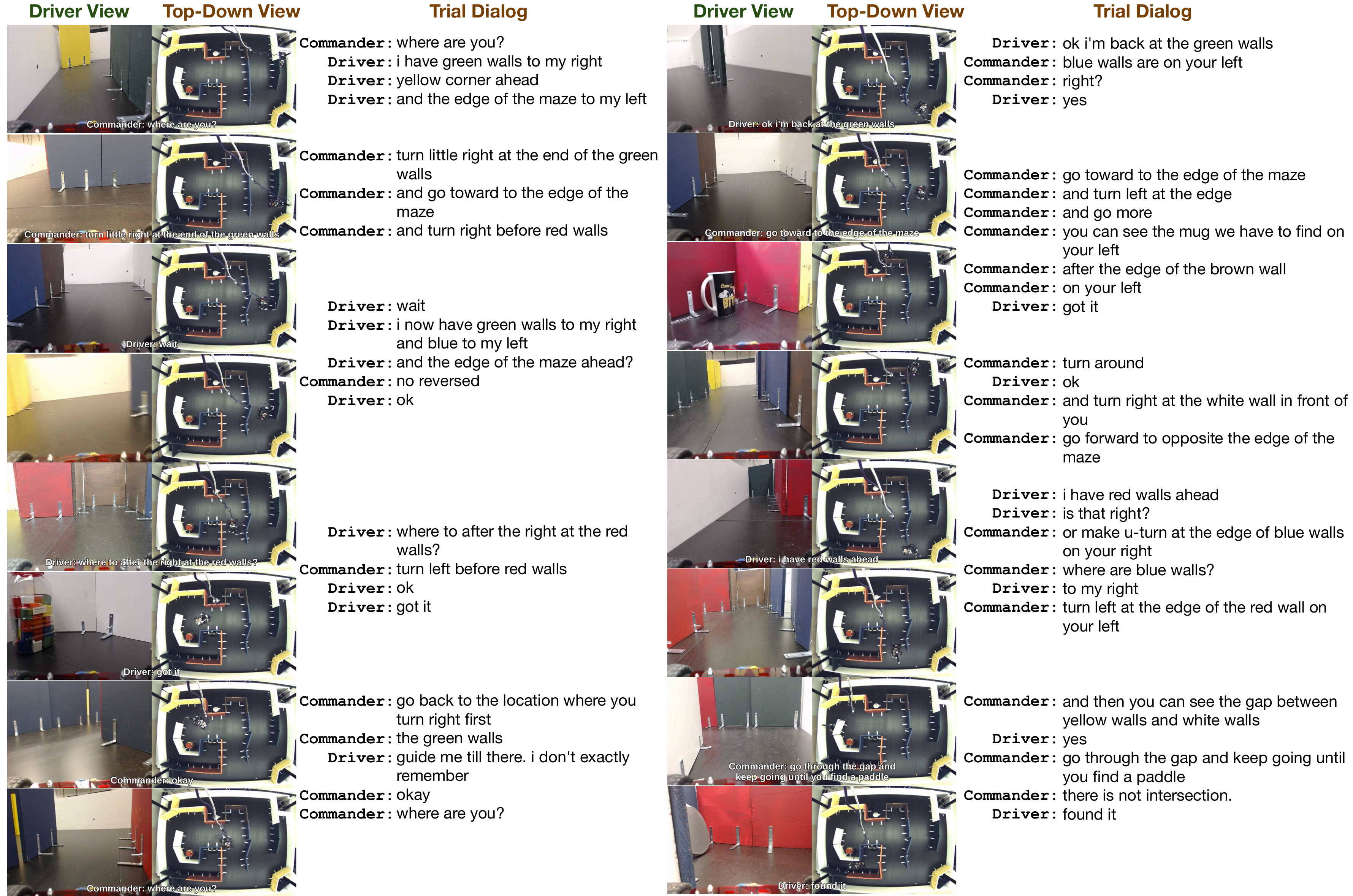}
    \caption{This example \dataset\ trial exhibits several phenomena analyzed in Section~\ref{ssec:analysis}.
    The \com\ uses commands like \textquoteinline{go back to the location where you turn right first} that need dialog history context to interpret.
    Additionally, communication repair and re-localization repair happen when the \dri\ suspects they have gone off-course.
    This arises twice, starting with \textquoteinline{wait $|$ i now have green walls to my right and blue to my left} and \textquoteinline{i have red walls ahead $|$ is that right?}
    }
    \label{fig:trial_353}
\end{figure*}

Figures~\ref{fig:trial_178}~and~\ref{fig:trial_353} summarize two full \dataset\ trials.
Trials each contain front-facing camera feeds, top-down camera feeds, and text dialog between the \com\ and \dri.
The demonstration video includes full-length recordings of the example trials shown here.

\subsection{Replay-Simulation Environment Details}

\begin{figure*}[t]
    \centering
    \includegraphics[width=1\textwidth]{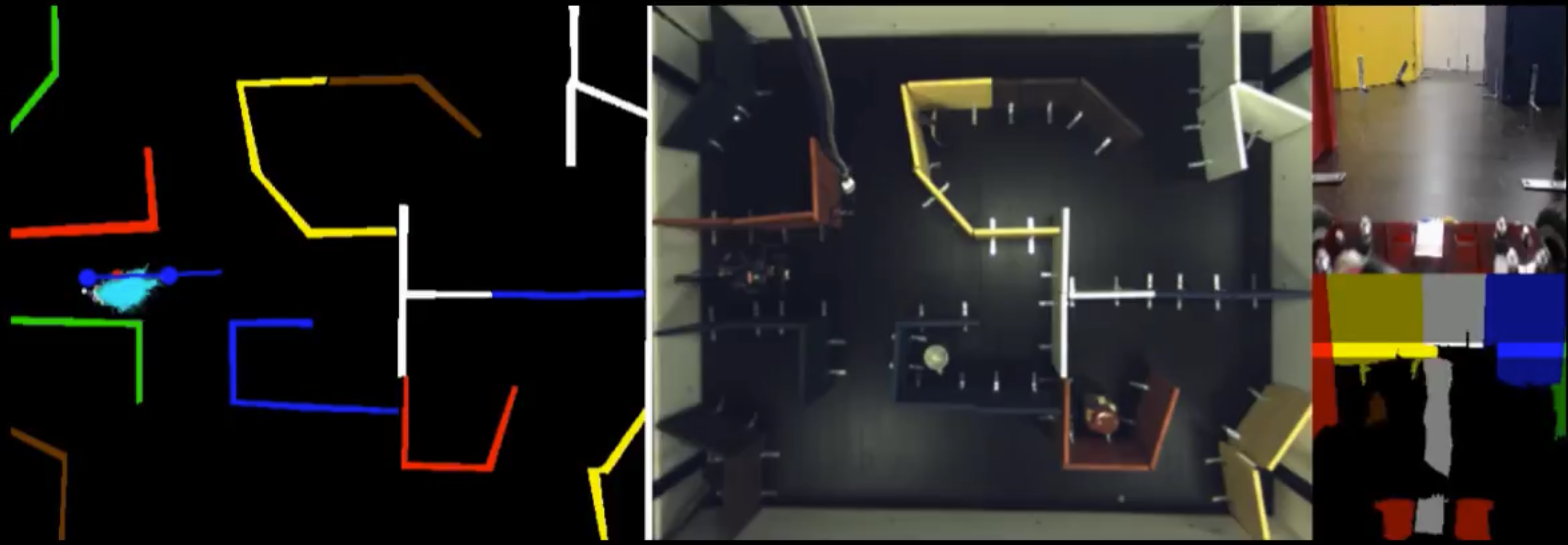}
    \caption{On the left is the annotated map used in a simulation. Blue dots represent particles from the particle filter. In the center is a top down view of the maze where the robot can be seen. In the upper right we see the \dri\ feed. In the lower right we see \dri\ feed separated in to its constituent colors. Particle filtering uses the constituent separated measurement to localize the \dri\ position successfully as shown by the clustering of the particles. The particle filter can be seen in action in the attached video.
    }
    \label{fig:replay_sim_breakdown}
\end{figure*}

We further detail the hyper-parameters used in our replay simulation.
The average number of actions to reach the goal is  $75.47 \pm 22.01$.
We use this average to inform a maximum episode length of $120$, about two standard deviation above the mean.
After inferring $120$ actions without inferring \texttt{stop}, models are stopped automatically.

We represent the jar, mug and paddle objects with distinct color blobs of light blue, pink, and olive respectively (Figure \ref{fig:replay_sim}).
When trying to find the first object, we initialize the agent's position at the \dri's initial location.
For subsequent objects, we initialize the agent facing the previous object.
Many sub-dialogs for finding objects begin with phrases that encourage the \dri\ to change direction, like \textquoteinline{turn around} in Figure~\ref{fig:trial_178}.

To validate the effectiveness of our simulation, we annotate and separate the \dri\ feed in to its constituent colors and use this measurement to localize the \dri's position in the maze. 
More details can be found in Figure~\ref{fig:replay_sim_breakdown}.
The successful localization of the driver shows that transfer is possible between our simulated setup and the real world \dataset\ setup.

\subsection{Further Modeling Details}

Our code is built on the backbone of prior VLN works \cite{anderson:cvpr18,thomason:corl19}.
Following their conventions, we train on the training fold when testing on the validation fold, but expand training data to include both training and validation when evaluating on the held out test set.
For test set evaluation, we report model performance when the model has trained for the number of epochs at which it achieved the best performance against the validation fold.
That is, we treat the epoch to train to as a hyperparameter set by the validation data per model.

We use the following hyperparameters in all our sequence-to-sequence models: a batch size of 100, a token embedding size of 128, an action embedding size of 8, the LSTM's hidden state size of 128, and the max dialog history token length of 100 (the most recent tokens are used as input).

Follow prior work~\cite{anderson:cvpr18,thomason:corl19}, we explicitly zero out the possibility of choosing actions that would lead to collisions, both at training and testing time.
This choice assumes a a robot can either detect that it made a collision and recover its previous position, or that it can detect a collision is imminent and override a model's choice to continue on a collision course.

For our teacher forcing models we use a learning rate of $0.0001$. For our student-forcing models we use $0.001$. We empirically find that that the student-forcing models train better and faster using this higher learning rate.
We also zero out the agent's \texttt{stop} logit until it is at the destination.
This choice prevents preemptive \texttt{stop} selection during \textit{student}-forcing, which can slow down training immensely given that our episodes are longer by almost a factor of 6 than previous works (i.e., an average of 75 actions versus 6 actions in VLN~\cite{anderson:cvpr18}).

\subsection{Statistical Analysis of Results}
We ran each model with three random seeds and took a micro-average of performance per-trajectory, then compared model performance using paired $t$-tests.
We ran paired $t$-tests between the full models under teaching and student supervision, as well as between full models and their unimodal ablations.
Data pairs are \navtask\ instances, and we compare the topological distance remaining to the goal between pairs of such instances under different models.
We apply a Benjamini--Yekutieli procedure to control the false discovery rate from running multiple $t$-tests.
Because the tests are not all independent, but some are, we estimate $c(m)$ under an arbitrary dependence assumption as $c(m)~=~\sum_{i+1}^{m}{\frac{1}{i}}$, where $m$ is the number of tests run.
We choose a significance threshold of $\alpha<0.05$.
In addition to the unimodal ablation test results in Table~\ref{tab:results}, we find that student supervision statistically significantly outperforms teacher supervision for full models.

\subsection{Success Rate and SPL}
Table~\ref{tab:results_sr_spl} presents trained agent results under the Success Rate and Success-weighted Path Length metrics defined in prior work~\cite{anderson:onevaluation}.
As recommended, we define the success of an episode by whether the final robot position was within two times the length of the robot body.
The most striking difference between success rate and topological distance (Table~\ref{tab:results}) is that under student-forcing, the agent \textit{never} stops within the bounds of the target object.
Our intuition is that while the agents trained with student-forcing get closer on average to the goal, they may do so by being conservative in their approach towards it, while those trained with teacher-forcing commit to a path and get as close as possible to where they believe the object to be, at the expense of often missing it.

\begin{table}[t]
    \centering
    \begin{tabular}{lccarar}
    & \multicolumn{2}{c}{\textbf{Inputs}} & \multicolumn{2}{c}{\textbf{Validation Success ($\pmb{\uparrow}$)}} & \multicolumn{2}{c}{\textbf{Test Success ($\pmb{\uparrow}$)}} \\ 
    \textbf{Training} & Vis & Lang & SR & SPL & SR & SPL \\ 
    \toprule
    Teacher & \cblkmark & & .01 $\pm$ .11 & .01 $\pm$ .10 & .06 $\pm$ .25 & .05 $\pm$ .20 \\ 
    Teacher & & \cblkmark & .00 $\pm$ .00 & .00 $\pm$ .00 & .00 $\pm$ .00 & .00 $\pm$ .00 \\ 
    Teacher & \cblkmark & \cblkmark & .03 $\pm$ .16 & .02 $\pm$ .14 & .04 $\pm$ .19 & .03 $\pm$ .16 \\
    \midrule
    Student & \cblkmark & & .00 $\pm$ .00 & .00 $\pm$ .00 & .00 $\pm$ .00 & .00 $\pm$ .00 \\ 
    Student & & \cblkmark & .00 $\pm$ .00 & .00 $\pm$ .00 & .00 $\pm$ .00 & .00 $\pm$ .00 \\
    Student & \cblkmark & \cblkmark & .00 $\pm$ .00 & .00 $\pm$ .00 & .00 $\pm$ .00 & .00 $\pm$ .00 \\
    \bottomrule \\
    \end{tabular}
    \caption{S2S model Success Rate (SR) and Success Weighted by Path Length (SPL) results.
    }
    \label{tab:results_sr_spl}
\end{table}